# SOFTWARE AGING ANALYSIS OF WEB SERVER USING NEURAL NETWORKS

Ms. G.Sumathi[1] and Mr. R. Raju[2]

[1]M. Tech (Pursuing), Computer Science and Engineering, Sri Manakula Vinayagar Engineering College, Pondicherry, India
`swasthik36@gmail.com`
[2]Associate Professor, Department of Information Technology, Sri Manakula Vinayagar Engineering College, Pondicherry, India
`rajupdy@gmail.com`

## ABSTRACT

*Software aging is a phenomenon that refers to progressive performance degradation or transient failures or even crashes in long running software systems such as web servers. It mainly occurs due to the deterioration of operating system resource, fragmentation and numerical error accumulation. A primitive method to fight against software aging is software rejuvenation. Software rejuvenation is a proactive fault management technique aimed at cleaning up the system internal state to prevent the occurrence of more severe crash failures in the future. It involves occasionally stopping the running software, cleaning its internal state and restarting it. An optimized schedule for performing the software rejuvenation has to be derived in advance because a long running application could not be put down now and then as it may lead to waste of cost. This paper proposes a method to derive an accurate and optimized schedule for rejuvenation of a web server (Apache) by using Radial Basis Function (RBF) based Feed Forward Neural Network, a variant of Artificial Neural Networks (ANN). Aging indicators are obtained through experimental setup involving Apache web server and clients, which acts as input to the neural network model. This method is better than existing ones because usage of RBF leads to better accuracy and speed in convergence.*

## KEYWORDS

*Software aging, software rejuvenation, rejuvenation schedule, ANN & RBF*

## 1. INTRODUCTION

Software aging is a phenomenon that refers to progressive performance degradation or transient failures or even crashes in long running software systems such as web servers. It mainly occurs due to the deterioration of operating system resource, fragmentation and numerical error accumulation [1]. Unexpected downtime cost due to software aging is high mainly in ecommerce websites and safety/business-critical applications. Software aging injures the usability of the software system and brings inconvenience to the users. It mostly occurs due to the accumulation of runtime errors. Runtime errors are the resultant of residual software effects such as memory leaking and unreleased file locks. These residual defects are difficult to be unveiled in the testing phase because there are few observable errors during the in-house testing phase. Even if they are unveiled, practical experience shows that most of corresponding errors are transient in nature [6], and difficult to be localized and removed. Therefore, these residual defects must be tolerated by users during operational phase.

Thus, like in humans, aging in software also cannot be avoided. We can just prolong the aging process or can reduce the effect caused by aging. So, the only possible solution to fight against aging is to reset the software system and clean its runtime environment before severe aging occurs, thereby avoiding system crash. This method is called software rejuvenation [7]. Software rejuvenation is a proactive fault management technique aimed at cleaning up the system internal state to prevent the occurrence of more severe crash failures in the future. It can maintain the robustness of software systems and avoid unexpected system outages.

## 2. OBJECTIVE OF THE PAPER

Every long running application must face the process of aging in its life time. This phenomenon of software aging, if left unseen will cause failure or even lead to the crash of entire system. Since aging cannot be avoided, the only remedy available to prevent the system from failure and to bring it back to the robust state is Software Rejuvenation. The important objective that must be taken into consideration is the appropriate time when the rejuvenation has to be performed, as periodic rejuvenation leads to wastage of cost because long running application such as a web server could not be put down now and then. This paper aims to obtain an accurate and optimized schedule for rejuvenation of a web server namely Apache. To do so, the status of aging indicators, the parameters that denotes aging process, has to be forecasted. The values are forecasted using Radial Basis Function based Feed Forward Neural Network. The proposed model leads to better accuracy and speed in convergence rate than the existing analytical and statistical models as these models assume the underlying probability distribution used for scheduling, leading to poor accuracy.

## 3. RELATED WORKS

The existing works on software aging is comes under either model-based approach or measurement-based approach. In model-based approach, certain assumption such as the origin and result of software aging are made and based on these assumptions a mathematical model is built. This model is either analytical or stochastic. The existing researches on model-based studies involve the proposal of a three state stochastic model to determine the best time to restart a telecommunication system switch [7] and Garg et al. analysed software aging in a transaction system and proposed an analytical model to estimated the probability of losing an arriving transaction and the expected response time of a transaction [8]. Dohi et al. built semi-Markov reward process models of software aging to solve the same problem [9, 10]. All these above mentioned models suffer from a drawback that the mathematical assumptions cannot be easily validated in practice and the derived mathematical properties are not useful for software maintenance [3].

In measurement-based approach, the aging indicators are monitored and based on the collected data the robustness of the system is assessed by applying statistical regression techniques such as auto regression [1], threshold auto regression [4] and autoregressive moving average model [5], to the monitored data. Wavelet Network is also used to solve the same problem [15]. The rate of software aging is usually not constant and it depends upon the system workload which varies according to time. Thus, time series model fits well to predict the future resource usage. Shishiny et al. proposed an MLP based Neural Network model to analyse software aging and predict the resource usage [16].

This work intend to provide a better optimal schedule for rejuvenation with increased accuracy rate and speed in convergence than the existing work, by using RBF based Neural Network to analyse resource usage data collected on a typical long-running software system namely, a web server, to assess the suitability of RBF based Neural Network for the analysis of software aging.

## 4. EXPERIMENTAL SETUP

To obtain an optimized schedule for rejuvenation, the aging indicators required for forecasting has to be obtained. Many aging indicators are available. For experimental purpose, three aging indicators namely, response time of the web server, used swap space and free physical memory are taken into consideration. In order to obtain the values of these aging indicators, an experimental setup shown in Figure 1 is made. The platform is deployed with an Apache server, two clients, and a 100M switch. The three computers are connected via the switch to each other. Linux Fedora 10 operating system is installed on the three computers, whose configuration is as follows:

Server configuration:
Processor Frequency: Dual Core 2GHz;
Memory: 2GMB;
Client configuration:
Processor Frequency: single-core 1.4GHz;
Memory: 512MB;

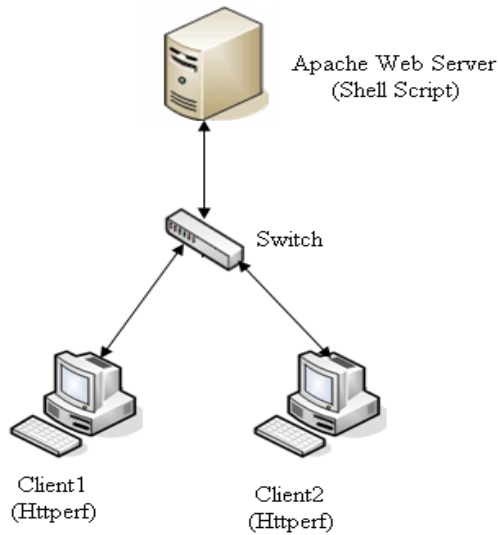

Figure 1. Experimental Setup

Httperf [17], a Web server test tool, is deployed on the clients to generate artificial concurrent requests with exponential time intervals to access the static html files on the Apache server. Since Apache has been well tested in practice, it is difficult for us to observe its aging symptoms in a short period under a normal runtime environment and the default parameter settings. It is necessary to find some way to expedite the aging of Apache. In the experiments, we adjust two parameters that are related to the accumulation of the effects of software errors:

MaxRequestPerChild and MaxSpareServers [2]. The first parameter limits the number of requests handled by each child process of Apache. The second parameter, MaxSpareServers, sets the maximum number of idle child processes. When the number of requests is low, some of existing child processes may be at idle state. If there are more than MaxSpareServers idle processes, Apache will kill excess ones. By setting it to zero, we can turn off this mechanism so that no child processes will be killed during runtime.

The used swap space and free physical memory are collected from /proc file system of Linux. From the /proc file system and with the help of the Linux monitoring tool procmon, measurements were periodically collected. Each Httperf request accesses one of five specified files of sizes 500 bytes, 5 kB, 50 kB, 500 kB, and 5MB on the server. Httperf is not only a workload generator, but it can also be employed for monitoring performance information. For collecting resource usage data over a long time period, a shell program was used to run httperf periodically. As for the connection rate, a value of 400 requests per second was chosen, which puts the web server in an overload state, and should speed up software aging.

## 5. ARTIFICIAL NEURAL NETWORK

Artificial Neural Network is one of the machine learning technique that is inspired by the organization and functioning of biological neurons. ANN has several advantages over statistical methods. Artificial neural network can be universal function approximators for even non-linear functions. It can also estimate piece-wise approximations of functions and has the ability to discover patterns adaptively from the data. When an appropriate number of nonlinear processing unit is given as input, neural networks can learn from experience and estimate any complex functional relationship with high accuracy [11]. Numerous successful ANN applications have been reported in the literature in a variety of fields including pattern recognition and forecasting [12].

### 5.1. ANN for Time Series Forecasting

The usage of ANN for time series analysis relies entirely on the data that were observed and is powerful enough to represent any form of time series. ANN can learn even in the case of noisy data and can represent nonlinear time series. For example, given a series of values of the variable x at time step t and at past time steps $x(t), x(t-1), x(t-2),…, x(t-m)$, we look for an unknown function F such that; $x(t+n)=F[x(t), x(t-1), x(t-2),…, x(t-m)]$, which gives an n-step predictor of order m for the quantity x.

## 6. RBFNN FOR SOFTWARE AGING FORECASTING

Radial Basis Functions emerged as a variant of ANN in late 80's. RBF's are embedded in a two layer neural network, where each hidden unit implements a radial activated function. The output units implement a weighted sum of hidden unit outputs. Unlike MLP network model, RBFNN does not have weight adjustment in the link between the input layer and hidden layer. In RBFNN weights are adjusted only in the link between the hidden layer and output layer. RBF networks have excellent approximation capabilities [18]. Due to their nonlinear approximation properties, RBF networks are able to model complex mappings, which perceptron neural networks can only model by means of multiple intermediary layers.

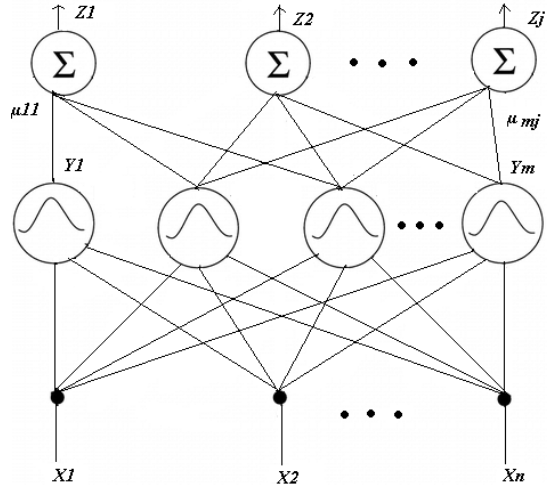

Figure 2. RBF Network Model

Figure 2 shows the radial basis function neural network. The bell shaped curves in the hidden nodes indicate that each hidden layer node represents a bell shaped radial basis function that is centered on a vector in the feature space. There are no weights on the lines from the input nodes to the hidden nodes. The input vector is fed to each m-th hidden node with the following radial basis function

$$Y_m = f_m(X) = \exp[-|X-C_m|^2/(2\sigma^2)] \quad (1)$$

where $|X-C_m|^2$ is the square of the distance between the input feature vector X and the center vector $C_m$ for that radial basis function. The values $\{Y_m\}$ are the outputs from the radial basis functions. The values equidistant from the center in all directions have the same values, so this is why these are called radial basis functions. The outputs from the hidden layer nodes are weighted by the weights on the lines and the weighted sum is computed at each j-th output node as

$$Z_j = (1/M)\sum_{(m-1, M)} \mu_{mj} Y_m \quad (2)$$

The collected data set is divided into two segments, one to train the RBFNN and the other for testing. The testing segment is used to evaluate the forecasting performance of the RBFNN in predicting the performance parameters' values, since the work proposed follows supervised learning technique.

### 6.1 Training the RBFNN

In order to predict the status of aging indicators to obtain an optimized schedule for rejuvenation, the RBFNN has to be trained such that the mean square error is minimized to the maximum extent. The mean square error function that is to be minimized by adjusting the parameters $\{\mu_{mj}\}$ is similar to the one for BPNN except that this is much simpler to minimize. There is only one set of parameters instead of two as was the case for BPNNs. Upon suppressing the index q we have

$$E = (1/J)\sum_{(j-1,J)} (t_j - Z_j)^2$$
$$E = (1/J)\sum_{(j-1,J)} (t_j - (1/M)\sum_{(m-1,M)} \mu_{mj} Y_m)^2 \quad (3)$$

Thus,

$$\partial E/\partial \mu_{MJ} = (\partial E/\partial Z_J)(\partial Z_J/\partial \mu_{MJ})$$

$$\partial E/\partial \mu_{MJ} = [(-2/J) \sum_{(J-1,J)} (T_J - Z_J)](Y_M/M) \quad (4)$$

Upon putting this into the steepest descent method

$$\mu_{mj}^{(k+1)} = \mu_{mj}^{(k)} + [2\eta/(JM)] \sum_{(j-1,J)} (t_j - Z_j)] Y_m \quad (5)$$

where $\eta$ is the learning rate, or step size, as before. Upon training over all Q feature vector inputs and their corresponding target output vectors, equation (5) becomes

$$\mu_{mj}^{(k+1)} = \mu_{mj}^{(k)} + [2\eta/(JM)] \sum_{(q-1,Q)} \sum_{(j-1,J)} (t_j^{(q)} - Z_j^{(q)})] Y_m^q \quad (6)$$

The center vectors $\{C^{(m)}: m=1,\ldots,M\}$ on which to center the radial basis function, the exempler vectors $\{X^{(q)}: q=1,\ldots,Q\}$ are considered as centers by putting $C^{(m)}=X^{(q)}$ for $m=1,\ldots,Q$. Once the network is trained, it is tested with the help of last segment of the observed data.

## 7. PARAMETERS FOR PERFORMANCE EVALUATION

The accuracy in forecasting the exhaustion of resources to obtain an optimized schedule for rejuvenation of web server is measured by Root Mean Square Error (RMSE) and Mean Absolute Percent Error (MAPE).

### 7.1 Root Mean Square Error

The root mean squared error, $E_i$ of an individual program $i$ is evaluated by the equation:

$$E_i = \sqrt{\frac{1}{n} \sum_{j=1}^{n} (P_{(ij)} - T_j)^2}$$

where $P_{(ij)}$ is the value predicted by the individual program $i$ for sample case $j$ (out of $n$ sample cases); and $T_j$ is the target value for sample case $j$.

For a perfect fit, $P_{(ij)} = T_j$ and $E_i = 0$. So, the $E_i$ index ranges from 0 to infinity, with 0 corresponding to the ideal.

### 7.2 Mean Absolute Percent Error

MAPE is calculated by averaging the percentage difference between the fitted (forecast) line and the original data:

$$\text{MAPE} = 1/n \left[ \sum_t | e_t/y_t | * 100 \right]$$

where $y$ represents the original series and $e$ the original series minus the forecast, and $n$ the number of observations.

16

## 8. RESULTS

The results produced by the RBF neural network model to predict an optimized schedule for rejuvenation of Apache web server are represented below.

### 8.1 Response Time

Figure 3 shows the graph of preprocessed response time of Apache web server collected for a period of about a week. The vertical line in the graph partitions the data into the set used for training and the set used for testing.

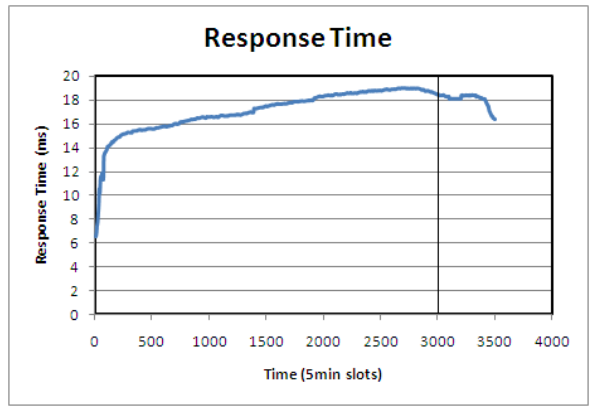

Figure 3. Response Time

Figure 4 shows the predicted response time obtained from the RBFNN model. The predicted results are compared with observed data segment. From the graph it is clear that the convergence obtained is faster and accurate.

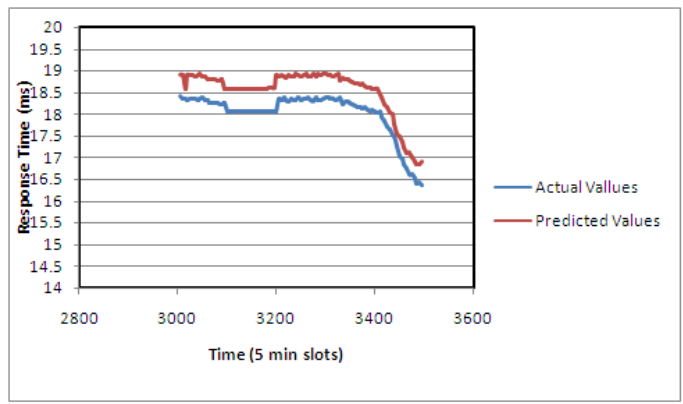

Figure 4. Predicted Response Time

### 8.2 Swap Space Used

Figure 5 shows swap space usage of the Apache Web server collected for a period of about 15 days. The graph shows considerable increases in used swap space at fixed intervals.

17

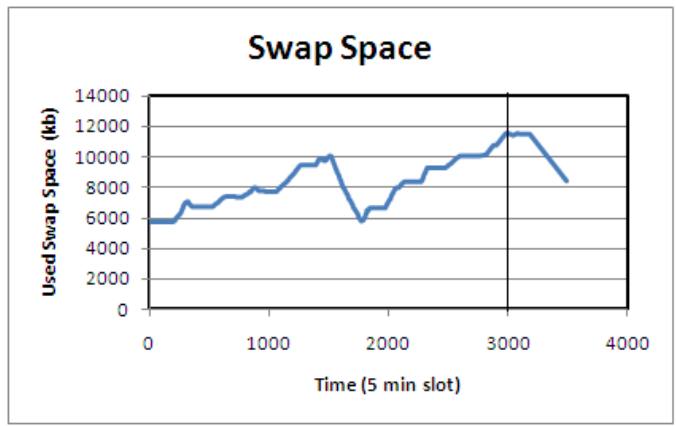

Figure 5. Swap Space Usage

Figure 6 shows the predicted swap space usage obtained by the RBFNN model. The predicted results are compared with the last segment of observed data used for testing which shows accuracy in prediction.

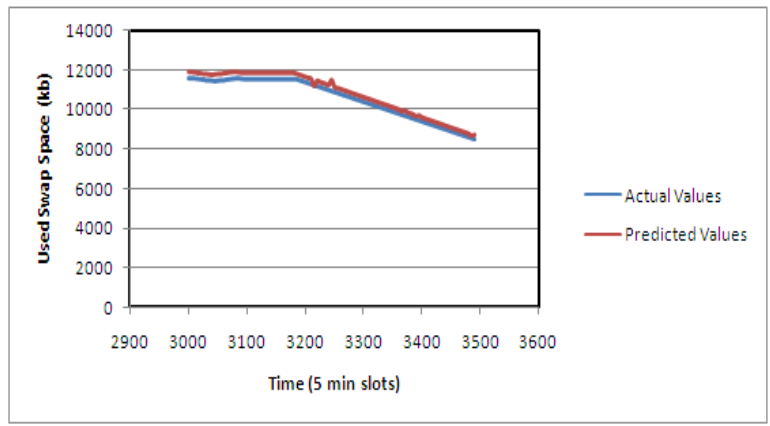

Figure 6. Predicted Swap Space Usage

## 8.3 Free Physical Memory

Figure 7 shows the graph of free physical memory of Apache web server collected for a period of about a week. The data is preprocessed by scaling because when the physical memory approaches it's allowed lower limit, the system frees up memory by paging. The vertical solid line in the graph segments the data into the set used for training and the set used for testing.

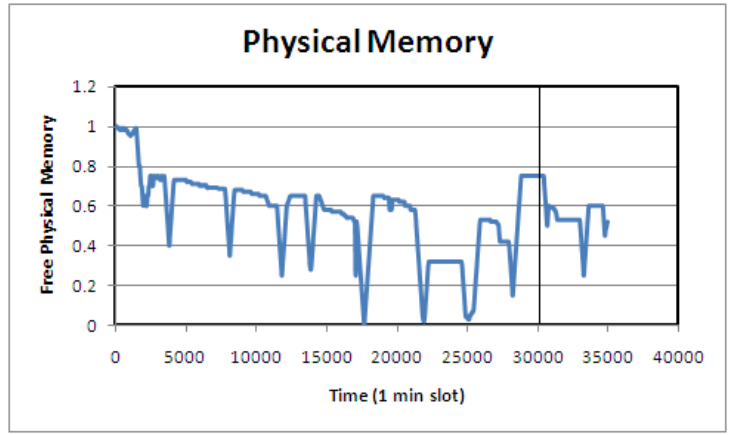

Figure 7. Free Physical Memory

Figure 8 shows the free physical memory space predicted by the RBFNN model. The predicted results are compared with the last segment of observed data set. From the graph it is clear that the convergence obtained is faster and accurate.

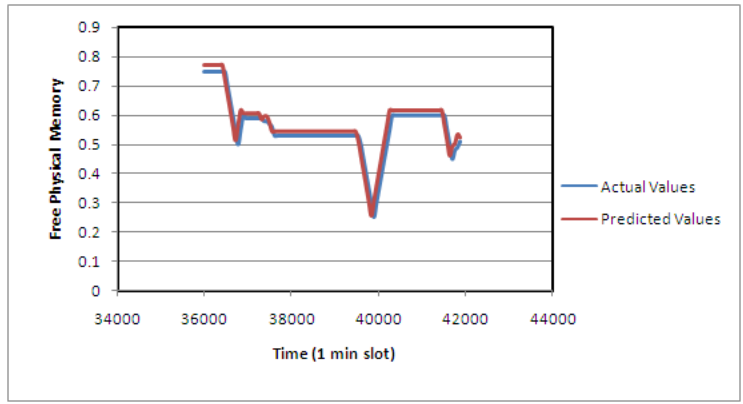

Figure 8. Predicted Free Physical Memory

## 9. EVALUATION OF PREDICTED RESULTS

The result of aging indicators predicted using RBF based feed forward neural network model is evaluated using the parameters RMSE and MAPE as shown in Table.1 and the results are compared with the existing MLP model.

Table 1. Evaluation of Forecasted Results

| Evaluation Parameters | MLP | RBFNN |
|---|---|---|
| RMSE | 0.7638 | 0.2546 |
| MAPE | 1.0935 | 0.4129 |

19

## 10. CONCLUSION

This paper worked on obtaining an accurate and optimized schedule for rejuvenating a web server (Apache web server for example) to prevent it from crashes due to the phenomenon called software aging.To obtain the schedule for rejuvenation the aging indicators are selected and real time values are obtained through experimental set up. A segment of the obtained values are used for training the RBFNN model which further predicts the expected results of aging indicators. The predicted results are evaluated using the accuracy parameters such as RMSE and MAPE. Also the obtained results are represented graphically to illustrate convergence rate and accuracy.

## 11. REFERENCES

[1] Michael Grottke, Lei Li, Kalyanaraman Vaidyanathan, and Kishor S. Trivedi, *"Analysis of Software Aging in a Web Server",* IEEE Transactions on Reliability, vol. 55, no. 3, September 2006

[2] Yun-Fei Jia, Lei Zhao and Kai-Yuan Cai, "*A Nonlinear Approach to Modeling of Software Aging in a Web Server",* 15th Asia-Pacific Software Engineering Conference, 2008

[3] Yun-Fei Jia, Jing-Ya Su, and Kai-Yuan Cai, *"A Feedback Control Approach for Software Rejuvenation in a Web Server"*, 978-1-4244-3417-6/08, 2008 IEEE

[4] Xiu-E Chen, Quan Quan, Yun-Fei Jia and Kai-Yuan Cai, *"A Threshold Autoregressive Model for Software Aging"*, Proceedings of the Second IEEE International Symposium on Service-Oriented System Engineering 0-7695-2726-4/06, 2006

[5] Lei Li, Kalyanaraman Vaidyanathan and Kishor S. Trivedi, *"An Approach for Estimation of Software Aging in a Web Server*", Proceedings of the 2002 International Symposium on Empirical Software Engineering 0-7695-1796-X/02, 2002

[6] A. T. Tai, L. Alkalaj, and S. N. Chau, *"On-board preventive maintenance: a design-oriented analytic study for long-life applications"*, Performance Evaluation, 35, 215–232, 1998

[7] Y. Huang, C. Kintala, N. Kolettis, and N. Fulton, *"Software Rejuvenation: Analysis, Module and Applications"*, in Proceedings of the 25th IEEE International Symposium on Fault-Tolerant Computing, pp. 381-390, Pasadena, USA, June 1995.

[8] S. Garg, A. Puliafito, M. Telek, and K. S. Trivedi, *"Analysis of Software Rejuvenation Using Markov Regenerative Stochastic Petri Net"*, in Proceedings of the Sixth International Symposium on Software Reliability Engineering, pp. 24-27, 1995.

[9] T. Dohi, K. Goseva-Popstojanova, and K. S. Trivedi, *"Analysis of software cost models with rejuvenation"*, in Proceedings of the International Symposium on High Assurance Systems Engineering, pp. 25–34, 2000

[10] T. Dohi, K. Goseva-Popstojanova, and K. S. Trivedi, "Estimating software rejuvenation schedules in high assurance systems", Computer Journal, 44(6):473–485, 2001

[11] Xin Yao, Senior Member, IEEE, *"Evolving Artificial Neural Networks"*, Proceedings of the IEEE, vol. 87, no. 9, September 1999

[12] Hornik, K., Stinchcombe, M., White, H., *"Multilayer* feedforward networks are universal approximators", Neural Networks 3, 551-560, 1989

[13] Zhang, G. Peter and Qi, Min, *"Neural network forecasting for seasonal and trend time series"*, European Journal of Operational Research 160, 501-514, 2005

[14] Hassoun,M. H., *"Fundamentals of Artificial Neural Networks"*, MIT Press, 1995

[15] Xu, J., You, J. and Zhang,K., *"A Neural-Wavelet based Methodology for software Aging Forecasting"*, IEEE International Conference on Systems, Man and Cybernetics, Volume 1, Issue , 10-12 Oct. 2005 Page(s): 59 – 63 Vol. 1, 2005

[16] Hisham El-Shishiny, Sally Sobhy Deraz, Omar B. Badreddin, *"Mining Software Aging: A Neural Network Approach"*, 978-1-4244-2703-1/08, 2008 IEEE

[17] D. Mosberger and T. Jin, *"Httperf - A Tool for Measuring Web Server Performance"*, in the First Workshop on Internet Server Performance, Madison, USA, June 1998.

[18] Park. J, Sandberg. J.W, *"Universal Approximation using Radial Basis Functions Network"*, Neural Computation, vol.3, pp. 246-257

[19] David Lorge Parnas, *"Software Aging"*, 0270-5257/9 4000 1994 IEEE

[20] Michael Grottke, Rivalino Matias Jr., Kishor S. Trivedi, *"The Fundamentals of Software Aging"*, 1st International Workshop on Software Aging and Rejuvenation, IEEE, 2008

[21] QingE WU, ZhenYu Han, TianSong Guo, *"Application of an Uncertain Reasoning Approach to Software Aging Detection"*, Fifth International Joint Conference onINC, IMS and IDC, 2009

## Authors

G. Sumathi is pursuing her M.Tech degree in Computer Science and Engineering in Sri Manakula Vinayagar Engineering College, Pondicherry University, Puducherry, India. Her research area includes Artificial Intelligence, Software Engineering and MANET. She presented papers in various International Conferences on Aritificial Intelligence and Networking.

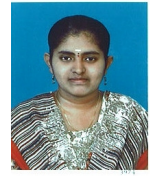

Mr.R.Raju is pursuing his PhD and received the M.Tech Degree in Computer Science and Engineering from Pondicherry University, Puducherry. His current research involves in Software Engineering and Cloud computing. He is presently working with Sri Manakula Vinayagar Engineering College (Affiliated to Pondicherry University, Puducherry) as Associate Professor in the Department of Information Technology.

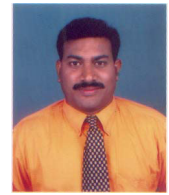